\documentclass[letterpaper]{article}
\usepackage{aaai23} 
\usepackage{times}  
\usepackage{helvet} 
\usepackage{courier}  
\usepackage[hyphens]{url}  
\usepackage{graphicx} 
\usepackage{natbib}  
\usepackage{caption}
\usepackage{algorithm}
\usepackage{algorithmic}

\usepackage{newfloat}
\usepackage{listings}
\usepackage{bm}
\DeclareCaptionStyle{ruled}{labelfont=normalfont,labelsep=colon,strut=off} 
\floatstyle{ruled}
\newfloat{listing}{tb}{lst}{}
\floatname{listing}{Listing}

\title{Thespian: Multi-Character Text Role-Playing Game Agents}
\author{
    Christopher Cui,
    Xiangyu Peng,
    Mark Riedl
}
\affiliations{
    Georgia Institute of Technology\\

    \{ccui46, xpeng62, riedl\}@gatech.edu

}

\usepackage{bibentry}

\usepackage{tikz}
\usepackage{enumitem}

\usepackage{graphicx,calc}
\newlength\myheight
\newlength\mydepth
\settototalheight\myheight{Xygp}
\usepackage{amssymb}

\usepackage{amsmath}
\usepackage{tabularx}
\usepackage[normalem]{ulem}
\usepackage{latexsym}
\usepackage{multirow}
\usepackage{booktabs}
\usepackage[normalem]{ulem}

\usepackage{enumitem}

\begin{document}
\nocopyright

\maketitle

\begin{abstract}
Text-adventure games and text role-playing games are grand challenges for reinforcement learning game playing agents. Text role-playing games are open-ended environments where an agent must faithfully play a particular character. We consider the distinction between {\em characters} and {\em actors}, where an actor agent has the ability to play multiple characters. We present a framework we call a {\em thespian agent} that can learn to emulate multiple characters along with a soft prompt that can be used to direct it as to which character to play at any time. We further describe an attention mechanism that allows the agent to learn new characters that are based on previously learned characters in a few-shot fashion. We show that our agent outperforms the state of the art agent framework in multi-character learning and few-shot learning.
\end{abstract}

\section{Introduction}
Text adventure games are those in which a player can only interact with an interactive environment through reading text descriptions of the environment and acting by typing descriptions of actions. Text games present a grand challenge for AI because they 
(a)~are partially observable; 
(b)~have combinatorially large state spaces consisting of all possible descriptive text strings;
(c)~have combinatorially large action spaces in the order of billions of possible text commands;
(d)~require reasoning about long-horizon causal dependencies;
and
(e)~require commonsense and narrative trope reasoning
\cite{hausknecht2020interactive}. Text adventure game playing has become a benchmark challenge for reinforcement Learning (RL) agents ~\cite{hausknecht2020interactive, narasimhan2015language, ammanabrolu2019playing, ammanabrolu2020graph, ammanabrolu2020avoid, adhikari2020learning}, which play by exploring the environment and receiving score based on how far they make it through the game.

\begin{figure}[!tbh]
\centering
 \includegraphics[width=0.9\linewidth]{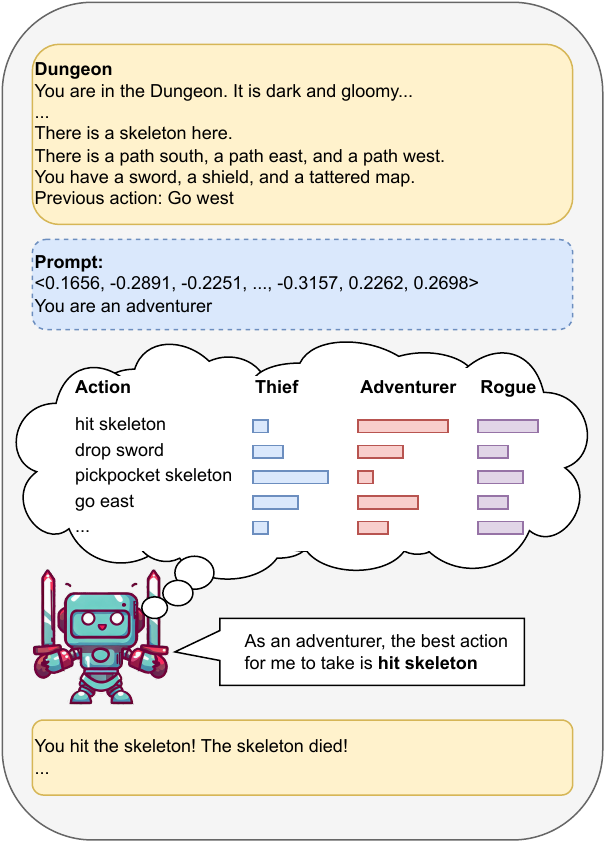}
\caption{A Thespian Agent is capable of acting out a number of different characters by being provided a prompt that indicates which character it should emulate at the time.}
    \label{fig:first}
\end{figure}

Relatedly, table-top role playing games, such as Dungeons \& Dragons, involve multiple players that interact with textual descriptions of the environment as well as dialogue with other players. 
While players may be motivated by a quest or mission, table-top role playing games are fundamentally {\em open-ended}, meaning that players can interact with the environment and with each other in ways that are not strictly dictated by a quest, mission, or set of puzzles.
Open-ended role-playing extends the same challenges of text adventure games but removes the environmentally-dictated reward structure.
The predominant question for open-ended role-playing is whether an agent acts consistently with a given character definition. 

Because there may be no explicit reward associated with progression in open-ended role playing games, an agent must instead be trained to, at least, emulate particular character types such as ``thief'' or an ``adventurer'', each of which has different preferences for different actions depending on the situation.\footnote{This is a simplification of table-top role-playing games that can also feature distinct character personalities and back-stories.} 

In this paper we consider the distinction between {\em character} agents and {\em actor} agents. A {\em character} agent is trained to act like one specific character; for all intents and purposes it {\em is} that character and knows nothing else but how to be that character. In contrast, an {\em actor} agent has knowledge about how to play many different characters and can receive instruction from an external source (for example a movie director or a dungeon master) about which character type to play.Furthermore, an actor can leverage the character knowledge to learn to blend characters with only a small amount of additional practice (e.g., {\em few-shot learning}) without exhaustively re-training from scratch. We will refer to actor agents as {\em thespian agents} to distinguish between agents that learn to enact multiple characters from the actor-critic reinforcement learning architecture.

This paper considers two challenges. The first is to train a single reinforcement learning agent model that can switch between character types with a simple instruction. We present a new RL agent that can learn to emulate multiple characters simultaneously with an updated policy model that generates $|C|$ sets of action distributions, where $C$ is a set of character classes. The agent also learns a soft prompt that can later be provided as a cue to emulate a specific character.

The second challenge is to be able to train a thespian agent to learn new characters in a faction of the training time while maintaining performance in the previously trained characters. We achieve this by adding an attention mechanism to the outputs of the the thespian agent, which can choose which can learn how to blend the action probabilities of different characters, thus learning a new character and a new soft prompt.

To return to our character vs. actor metaphor, we now have an thespian model that can simultaneously generate different actions for different characters. This is equivalent to a thespian thinking about how different characters will respond to the same situation.
The thespian agent receives direction in the form of a prompt indicating what character to play. If the thespian needs to play a new character that it has never played before it can learn a new prompt for the new character much faster than if it had to learn from scratch by leveraging what it already knows about playing other characters.

We conduct experiments across two original character types, a ``thief'' and an ``adventurer'' and demonstrate the ability of a single thespian agent trained on both characters to perform as well as separate baseline models trained to emulate individual characters. We show that we can use a novel attention mechanism to learn a third character that is a blend of the previously trained characters in a few-shot fashion. This few-shot character learning is 10x faster than baseline alternatives and doesn't degrade the performance of original characters.

\section{Background and Related Work}

The distinction between characters and actors have been made before.
\citet{louchart2007synthetic} consider an actor agent one that makes a secondary assessment of its own cognitive and emotional state.
\citet{riedl2003actor} consider an actor agent one that doesn't just reason about the best action to convey a character but also incorporates directorial goals. \citet{si2005thespian} consider an actor agent one that reasons about the cognitive state of other interlocutors in an interactive game; they also referred to their agent as a {\em thespian}.
These prior works looked at acting as meta-cognition, but agents could not represent more than one character without retraining or reprogramming. While our work can also be considered a form of meta-cognition, our focus is on a single model trained to be able to reason about and enact different characters.

\subsection{Text Adventure Game Playing Agents}

Text adventures are games in which the player must read textual descriptions of the environment and describe their actions with short text commands. Most text adventure games have a narrative progression through puzzles toward an ultimate goal or conclusion. Text based games have shown great potential for use as Reinforcement Learning benchmark environments~\cite{hausknecht2020interactive, narasimhan2015language}. 
\citet{ammanabrolu2019playing} proposed augmenting reinforcement learning with knowledge graphs as external memory about world state.
\citet{ammanabrolu2020graph} proposed KG-A2C, which integrates knowledge graphs into the actor-critic~\cite{bahdanau2016actor} RL framework.
The Q*BERT agent~\cite{ammanabrolu2020avoid} further extended KG-A2C to incorporate the BERT~\cite{devlin-etal-2019-bert} language model into the model architecture. We build on top of the KG-A2C family of models since they have shown state-of-the-art performance. Other techniques for playing text-based games include GATA~\cite{adhikari2020learning}, which builds a knowledge-graph based  representation of the world on top of a transformer-based agent, training through a combination of RL and self-supervised learning.

\subsection{Text-based Role Playing Agents}

Whereas text adventure games have pre-defined progression toward a goal state, table-top role playing games involve open-ended game play.
We refer to text-based environments that support open-ended game play as {\em text-based role playing} to signify the interaction with the environment through reading and writing text instead of verbal interactions with other players and game masters.

The LIGHT environment~\cite{urbanek2019learning} is a crowdsourced text-based role playing game with a rich environment with interactable NPCs, objects and locations, each with a short paragraph description, demonstrating the value of grounding in training agents that can not only act but also converse successfully. \citet{ammanabrolu2021motivate} propose agents that can switch seamlessly between generating natural language and action declarations. These agents can learn to play different characters when given a motivation that includes character type and goal as part of the input world state. This work is most similar to ours, except our agents do not require explicit motivations or goals beyond a learned character prompt.

{\em Story Shaping}~\cite{peng2023story} is a technique for training RL agents to play text role-playing games wherein a story is converted into a rich reward signal. The technique can be used to train different characters, but can only train a single agent to emulate a single character. Our character-based reward strategy is related, but our rewards are manually crafted instead of inferred from stories.

\subsection{Few-Shot Adaptation}

Large pre-trained Language models have emerged as extremely powerful tools for NLP tasks\cite{devlin2019bert, raffel2020exploring, brown2020language}. However, a limitation of these powerful models is their size, some with parameters numbering in the billions \cite{brown2020language}. This makes them prohibitively expensive when it comes to further training or fine-tuning. Low-Rank Adaptation (LoRA) circumvents this by keeping the model frozen and introducing trainable rank decomposition matrices. Our proposed technique also freezes the core model and trains additional layers on top, though the specific mechanics needed for reinforcement learning are different.

Prompt-tuning also avoids the need to do further training on the model itself by introducing trainable, soft prompts that learn an ideal input based on the desired output~\cite{lester2021power}. \cite{peng2022model} proposes pairing soft prompts with an attention module to induce language models to perform different tasks. Using knowledge from a previously trained task to improve learning on a new task has also been explored by \cite{zhao2021meta}, their approach more focused on generalization across simpler objectives and adaptation to unseen environments.

\section{Preliminaries}

\subsection{Textworlds as RL Testbeds}

A text-adventure or text-based role playing game can be modeled as a partially-observable Markov decision process (POMDP) M = $\langle S, T, A, \omega, O, R, \gamma \rangle$ where
$S$ is the set of ground truth world states,
$A$ is the set of actions,
$T$ is the probability of transitioning from one state to another given an executed action,
$R$ is a reward function,
$O$ is the set of possible observations,
$\omega$ is the probability of observations given the ground truth world state, and $\gamma$ is a parameter estimating the reward horizon~\cite{hausknecht2020interactive}. In our setting, we will use a deterministic transition function $T$, which is common in text-based games. However, nothing in our proposed technique strictly requires it.
The objective of reinforcement learning is to learn a policy, $\pi:S \rightarrow A$ that maps states to actions, such that taking the action mapped to the current state and following the policy henceforth maximizes expected reward. 

\subsection{LIGHT} 

Our agent is trained in the LIGHT environment~\cite{urbanek2019learning}, a text world environment with a database of 1775 Non-Player Characters (NPCs), 663 locations, and 3462 objects with rich text descriptions. Game maps can also be handcraft with specifically placed NPCs, locations and objects. We create a map for our experiments such that multiple character types can have relevant activities to perform, including interacting with objects and NPCs.  
For example there are dragons for an ``adventurer'' character to slay, and armor to don, whereas a ``thief'' character can take money from the donations receptacle in a sanctuary.

Our experiments use base character types of ``Thief'' and ``Adventurer''.  We also associate rewards to different actions for each character type. For example, a ``Thief'' character agent is rewarded for obtaining a hidden dagger, stealing, and other thief-like actions.
Likewise, an ``Adventurer'' character agent is rewarded for obtaining a sword and armor from the armory and killing monsters, and other adventurer-like actions. There is no requirement that an agent do particular actions and no prescribed order. This is equivalent to the {\em Story Shaping} technique\cite{peng2023story} , except the rewards are manual, which is done to make more controlled experiments. Regardless of character type, all games terminate when the agent enters a particular, preset ``goal room'', at which time the agent receives a final reward that is smaller than the others. The entire game map is provided in the appendix.

\subsection{KG-A2C}
\label{sec:kg-a2c}

We build off the KG-A2C agent framework~\cite{ammanabrolu2020graph}, an Advantage-Actor Critic architecture augmented with a knowledge-graph based attention. KG-A2C's space of observations includes (a)~text description of the room the agent is in via the ``look'' command,
(b)~text descriptions of the character's inventory via the ``inventory'' command,
(c)~the agent's last command, and
(d)~feedback from the last command.
The state observations are concatenated and embedded using a recurrent GRU.

Simultaneously, the state observation is used to update a knowledge graph of facts about the world that have been observed to date.
This includes facts and relations about rooms, objects in rooms, inventory items, etc.
This knowledge graph is then embedded using a graph attention mechanism~\cite{veličković2018graph}. 

Advantage-actor critic networks \cite{mnih2016asynchronous} have two heads. The actor head generates logit scores, one for each possible action, which can be converted to a probability distribution via softmax and sampled to determine which action the agent takes. The critic head estimates the utility of the state. Actions are made up of verbs and optional object names. The KG-A2C agent generates a verb, which maps to a pre-defined template, and the generated object name is used to populate the template. 

\section{The Thespian Agent}

\begin{figure*}[tbh!]
\includegraphics[width=\textwidth]{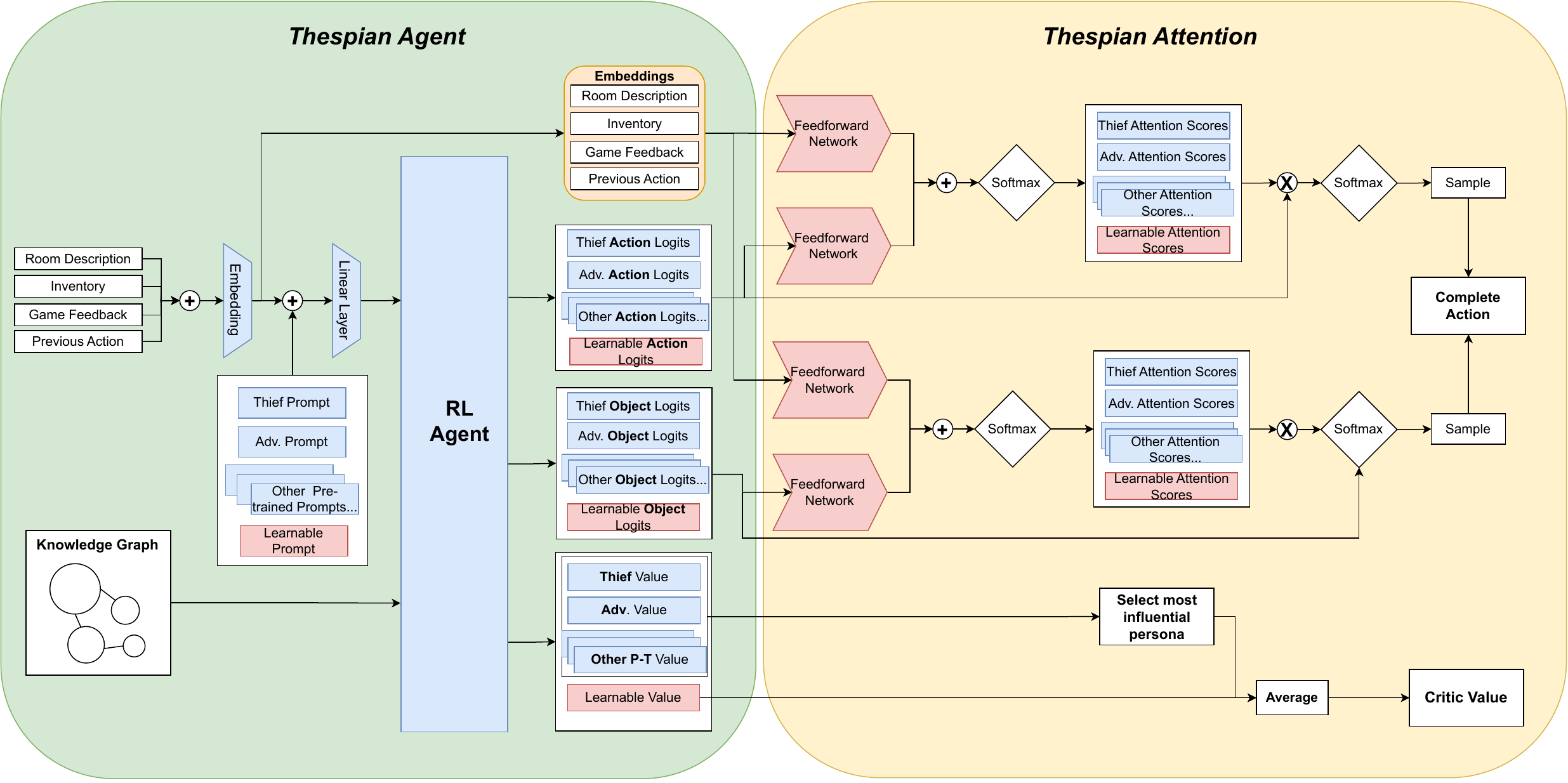}
    \caption{The Thespian Attention takes in the embedded observations and the stacked logits, calculating an attention score per character for each observation. When training the Thespian Attention, the blue-shaded boxes indicate frozen modules with red-shaded boxes being trainable modules.
    }
    \label{fig:weight_calcs}
\end{figure*}

Building off the basic framework of KG-A2C 
we describe how a single agent policy model can learn to emulate multiple characters. To train an single model to emulate different characters, it must be rewarded differently for each character, which can confuse an agent unless it has a way of disentangling the characters.
Our thespian agent architecture addresses this challenge in two ways. 
First, we provide a means to learn {\em soft character prompts}.
These are unique codes that are associated with different characters and can be provided as input to indicate which character the agent should emulate. Second, we change the actor and critic heads to generate sets of logit scores for all learned characters. Thus the agent can reason about which actions are best for each character, and we can sample from the set of logits for which ever character we want to execute. Figure~\ref{fig:weight_calcs}(left, green box) shows the thespian agent, focusing on the these two aspects.

\subsection{Character Prompts}

First, we allow for a {\em soft character prompt} to be learned. 
Each prompt is associated with different character the model has been trained to emulate and induces the agent to generate behavior that is consistent with the associated character. This is similar to the notion of the {\em soft prompt}\cite{lester2021power}, which is like a regular prompt for LLMs but given as an embedding instead of natural language. 
The soft character prompt vector of values can be interpreted as an instruction analogous to saying ``I am in state {\em x} and I am a Thief. My next action would be...'' at the embedding level.

Let
$\mathbf{P}=\bigl[\begin{smallmatrix} \mathbf{p}_1  \\ ...  \\ \mathbf{p}_n\end{smallmatrix}\bigr]$
be a set of soft character prompts for each character $c_i\in C$ and let $\mathbf{o}$ be the embedded current state observation. 
Initially, the prompts $\mathbf{p}_i$ are empty, initialized with random numbers. The internal state representation $s_i$ for character $c_i$ is:
\begin{equation}
\mathbf{s}_i = W^T_i \times \textit{cat}(\mathbf{o}, \mathbf{p}_i)
\end{equation}
where $W$ is a set of trainable weights.

The soft character prompts are learned as follows. During training, the agent will engage in reinforcement learning games as normal. In each game, the agent will be provided with a different reward function for each character $c_i$. That is, a thief will be rewarded for certain actions and an adventurer will be rewarded for different actions.
The character, corresponding character reward function, and character prompt $\mathrm{p}_i$ are rotated each game to balance the training of multiple characters. Over time, each soft prompt is updated via gradient flow through $W$ such that each unique prompt is associated with a particular way in which the agent is rewarded.

\subsection{Character-Specific Action Scores}

We also modify the agent model's actor and critic modules.
The standard A2C framework produces logit scores for each action.
This vector of logit scores is traditionally converted to a probability distribution with a softmax layer and sampled to determine which action the agent takes. Our thespian agent model instead produces a stack of action logit scores. A softmax over this stack of logits produces $n$ probability distributions, for $n$ characters.

The critic head is likewise modified to produce $n$ predicted utility scores, one for each character.

Thus, the agent is simultaneously determining which action is best for each character and how good the current state is from the perspective of each character.

At training time, the characters are rotated each game and the $i$th set of logit scores is sampled to determine the agent's action, and the $i$th utility value is used to compute character-specific advantage loss. The loss is backpropagated through only the logits and utility used.

\section{Thespian Agent Experiments}

In this section we evaluate the thespian agent without the additional few-shot learning attention mechanism in order to determine the extent to which the agent can learn more than one character at a time.
In this experiment we train a single agent to emulate two characters: thief and adventurer. We execute the agent in the same general environment that has multiple opportunities for thief-specific actions and adventurer-specific actions. The environment (see Figure~\ref{fig:full-map} in the Appendix) has a common starting room and an exit room that terminates the game when the agent enters it. 
There are a cluster of thief-specific and adventurer-specific rewards clustered near the starting room. The environment then branches with one branch heading to areas that only contain thief-specific rewards and another branch heading to areas that only contain adventurer-specific actions.

The thespian agent is trained as follows. We create empty prompts for thief and adventurer. We train on one character reward, accompanied by the character prompt, for two games, then switch to the net character reward and character prompt for two more games. A game completes when the agent navigates to the exit room as described in Section 3.2.
We train for a total of 10,000 games and use the checkpoint with the highest performance on 20 test game runs, split equally between each character.

We evaluate the agent in the same environment, executing the agent with with each character prompt one at a time. We measure the percentage of total character-specific action opportunities the agent takes. We run each character prompt for 100 games with different initialization seeds and take the average result.

We compare to a baseline KG-A2C trained with the same training method (but without the prompts since the base KG-A2C architecture would not understand them), as well as the thespian agent with a prompt made of random numbers.

Table~\ref{tab:general-agent} shows the results. The base KG-A2C when trained only on thief rewards or adventurer rewards is able to achieve most of the character-specific score. The base agent trained on one character rarely attempts to perform actions specific to another character, which is to be expected and demonstrates that the environment setting is fair if the objective were to only train one character at a time. However, when the base KG-A2C is trained with both character rewards, the agent's performance relative to both characters suffers.
The resulting agent also attempts to get all rewards, regardless of character, thus failing to differentiate between characters.

\begin{figure}[tb]
    \centering
    \includegraphics[width=\linewidth]{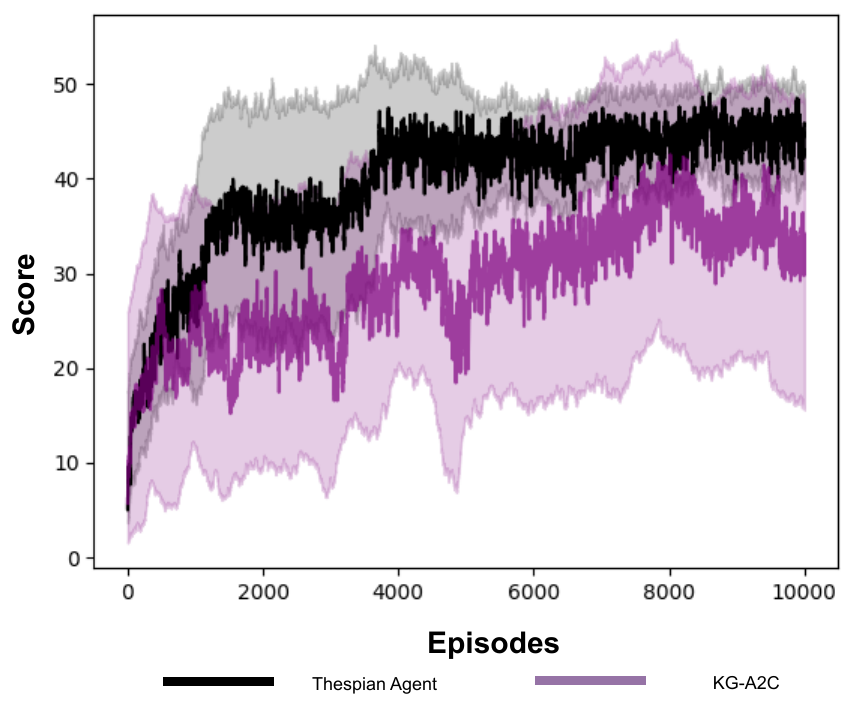}
    \caption{We see the thespian agent achieves convergence after about 4,000 episodes where the KG-A2C still struggles to perform even after 10,000 episodes.
    }
    \label{fig:t-vs-kga2c}
\end{figure}

In comparison, thespian agent uses a single model and that single model scores well has a high thief score when given the thief prompt and a high adventurer score when given the adventurer prompt. 
The thespian agent rarely attempts actions that are specific to a non-prompted character. Despite being trained on multiple character rewards, the thespian agent achieves performance equivalent to the base model trained on only one character. Figure~\ref{fig:t-vs-kga2c} shows the learning curve of the single thespian agent training on both characters versus a single base KG-A2C training on both characters using the same character rotation scheme. As can be seen KG-A2C gets trapped in a local maximum.

When the thespian agent is given a random prompt, it scores poorly as either character. There may be a bias in the environment that leads the agent to prefer the branch that contains more adventurer score, explaining why the agent obtains more adventurer rewards.

\begin{table}[tb]
\caption{Performance of 
the base KG-A2C trained in different conditions and a single thespian model responding to different prompts.
}~\label{tab:general-agent}
  \centering
\footnotesize
\setlength\tabcolsep{1pt}
\begin{tabular}{m{1.6cm}|m{1.0cm}m{1.0cm}|m{1.0cm}m{1.0cm}|m{0.8cm}}
\toprule
\textbf{Experiment} & \textbf{Thief score \%} & \textbf{Thief  std.dev}  & \textbf{Adv score \%} & \textbf{Adv  std.dev} & \textbf{Avg. Game Steps}\\ 
\midrule
\multicolumn{6}{c}{\bf Base KG-A2C}\\
\midrule
Thief-Only & 93.2 & 9.3 & 3.8 & 0 & 24.7 \\\midrule 
Adv-Only & 3.8 & 0 & 99.3 &  4.8 & 33.4\\ 
\midrule
Both trained & 88.2 & 10.3 & 68.6 & 7.8 & 33.1 \\
\midrule
\multicolumn{6}{c}{\bf Thespian agent}\\
\midrule
Thief Prompt & 92.9 & 17.4 & 3.8 & 0 & 22.1 \\ 
\midrule
Adv. Prompt & 4.3 & 0 & 99.1 & 11.1 & 32.2\\ 
\midrule
Rand. Prompt & 10.7 & 12.8 & 72.7 & 37.5 & 26.5 \\
\bottomrule
\end{tabular}

\end{table}

\section{Few-Shot Learning with Thespian Attention}
\label{sec:attention}

The thespian agent is a single agent that can be trained to emulate many different characters by providing one of the learned prompts as a cue for how to behave in an open-ended fashion.
In this section we consider the question of whether a pre-trained thespian agent can learn a new character that draws on knowledge of previously learned characters.

Given a thespian agent that has been trained on $n$ characters, training the $n+1$th character poses challenges. 
Training on the $n+1$th character, with a new reward runs the risk that the agent forgets the previous $n$ characters.
This is a commonly known phenomenon with fine-tuning any type of model.
It is typically a desired phenomenon when we wish to update the model to a new behavior that overwrites the pre-trained behavior. 
However, in this case, we wish to preserve the ability to execute previous behavior while adding new behavior. 

Our approach is to freeze the thespian agent model and add a module (see Figure~\ref{fig:weight_calcs} right, yellow) with learnable weights that operate on the original, frozen model's outputs.
Since we seek to teach the agent a new character that is a blend of existing characters, we apply an attention mechanism across the action logit scores for each character.
This attention module learns to blend the raw logit scores for each characters to produce a single final action probability distribution.

Specifically, we adapt the attention module from \citet{peng2022model}---which is used for few-shot learning in LLMs---to the reinforcement learning setting.\footnote{In place of the embedded token sequence, we use the embedded observation tensors $o_t$ but do not perform a maxpool over the embedded observations as they are much smaller than the token sequences used in Peng et all's model ensemble}

\subsection{Thespian Attention}

Let $\mathbf{O}=\biggl[\begin{smallmatrix} \mathbf{o}_\mathrm{look}  \\ \mathbf{o}_\mathrm{inv}   \\ \mathbf{o}_\mathrm{prev} \\ \mathbf{o}_\mathrm{fback}\end{smallmatrix}\biggr]$
represent the stacked observation component embeddings, which  
is fed through a feed-forward network, projecting it to a new, non-linear representation space,
\begin{equation}
\mathbf{h_O} = LN(W^T_\mathrm{FF2} \times \gamma(W^T_\mathrm{FF1} \times \mathbf{O}))
\label{eq:ff}
\end{equation}
with $\gamma$ as a non-linear activation function, $W_\mathrm{FF1}$ and $W_\mathrm{FF2}$ as trainable weights, and $LN(\cdot)$ is a Layer Norm~\cite{ba2016layer}.

The action logits $\mathbf{a}_i$ for all characters $c_i\in C$ produced by the frozen thespian agent are stacked as $\mathbf{A}=\bigl[\begin{smallmatrix} \mathbf{a}_1  \\ ...  \\ \mathbf{a}_n\end{smallmatrix}\bigr]$   
and also fed through a feed-forward network identical to Equation~\ref{eq:ff} to obtain $\mathbf{h_A}$. To obtain the final set of attention scores for each observation we perform a matrix multiplication between $\mathbf{h_O}$ and $\mathbf{h_A}$.
We divide by a constant $m$ that applies a temperature-like smoothing before applying a softmax layer to obtain the matrix of attention scores,
\begin{equation}
\mathbf{S} = \textit{softmax}\Bigl(\frac{\mathbf{h_O} \times \mathbf{h_A}}{m}\Bigr)
\end{equation}
with $c$ being some character  (pre-trained or being learned few-shot). 

The final weighted averaged logits for the action is:
\begin{equation}
\mathbf{p}_\mathrm{final} = \textit{softmax}\bigl( \mathbf{\alpha}_\mathrm{obs} \times \mathbf{S}^T \times \mathbf{A} \bigr)
\end{equation}
where $\mathbf{\alpha}_\mathrm{obs}$ is a vector of scaling coefficients for each of
$\mathbf{o}_\mathrm{look}$, $\mathbf{o}_\mathrm{inv}$,
$\mathbf{o}_\mathrm{prev}$, and $\mathbf{o}_\mathrm{fback}$,
the look, inventory, previous action, and previous action feedback components of the state observation, respectively.

$\mathbf{\alpha}_\mathrm{obs}$ is a hyperparameter that allows us increase the influence of different parts of the observation.
They can be equal and sum to one to have a uniform averaging effect, or be used to increase or decrease the contribution of each component of the state observation.
Setting the coefficients $>1.0$ loads greater probability mass onto the highest-scoring action score logits. 
This has the effect of making the agent more ``exploitative'' when sampling from the probability distribution over actions.

The result is that the thespian attention learns the optimal weights to calculate the contribution of each pre-trained character in determining an action for the new character in  the current state with respect to each observation tensor.

Since the KG-A2C base splits action generation into verb and object selection, the above process is repeated for the verb and the object to produce one probability distribution for the verb and one distribution for the object.
The sampled verb and sampled object are combined using the KG-A2C template approach described in Section~\ref{sec:kg-a2c}.

\subsection{Few-Shot Training}

The traditional actor-critic loss is computed as the difference between the agent's predicted value of an action and the true expected value. 
However, the thespian agent produces a real-numbered utility value prediction for each character. 
Rather than perform a weighted average with the attention scores as we did for the action logits, we take the average of the predicted values of the state from the new character's perspective and the predicted value of the {\em most influential} pre-trained character.
This is the pre-existing character that the agent thinks has the best chance of receiving reward even though the reward function is for a new character.
Thus loss is a function of how much better the thespian attention can pick an action for the new character over the best chance if it had to play a pre-existing character.

The thespian agent can now be trained as before, by providing a new character reward and an empty prompt.
With the core thespian agent weights frozen, the agent will retain the ability to respond to existing character prompts.
The thespian agent will learn new weights in the feed-forward networks that combine the existing characters action logits.
We no longer need to specify which set of character action logits to sample from.
It will also learn a new prompt for the new character.

\section{Few-shot Experiments}
\begin{figure*}[!tbh]
    \centering
\includegraphics[width=\linewidth]{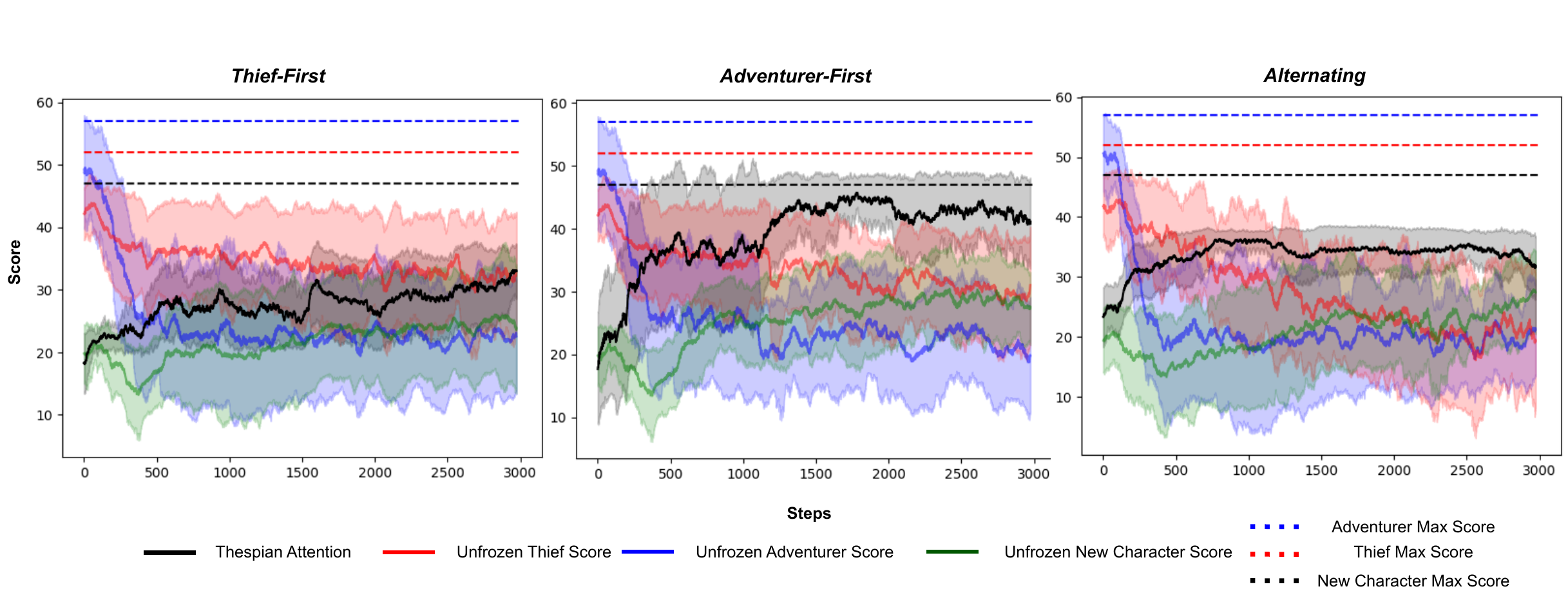}
    \caption{Scores of the thespian attention versus unfrozen agents for each game. The red and blue lines represent the score of the unfrozen agent on the ``Thief'' and ``Adventurer'' games when provided the respective prompt. We see that in all cases, the unfrozen agent's performance in the pre-trained characters suffers. In the \textbf{Adventurer-First} and \textbf{Alternating} maps, the thespian attention converge far sooner than the unfrozen agent. While not yet converged, we also see the thespian attention train faster than the unfrozen agent. 
    }
    \label{fig:Games}
\end{figure*}

The thespian attention uses far fewer parameters than the core agent.
Therefore we test the ability to train the thespian attention module to learn a new character in fewer training steps versus training from scratch.
Given a frozen thespian agent pre-trained to respond to the thief and adventurer prompts,
we train a new character---a ``Rogue''---that excels at both thieving and adventuring.
To demonstrate few-shot learning, 
we limit the total training steps to 3,000.

We created three variations of the environment:

\begin{itemize}
\item \textbf{Thief-first map}: 
all the thief-specific activities are arranged close to the starting room while all the adventurer-specific activities are closer to the exit room.

\item \textbf{Adventurer-first map}: 
all the adventurer-specific activities are arranged close to the starting room while all the thief-specific activities are closer to the exit room.

\item \textbf{Alternating map}:
the character-specific activities alternate between thief and adventurer as the agent progresses farther from the starting room.
\end{itemize}
These alternative maps demonstrate robustness to alternating conditions in the environment that require either knowledge about how to act as a thief or knowledge about how to act as an adventurer. For all characters, the maximum score the agent can achieve is $47$ and all characters share the same exit room. 
We use the total score achieved as a measure of how well the thespian attention allows the agent to learn a new character based on the pre-trained characters.

\subsection{Baselines}
\label{sec:baselines}

We compare two agents:
\begin{itemize}
\item \textbf{Thespian attention agent}: a pre-trained thespian agent with frozen weights and the few-shot attention mechanism.
\item \textbf{Unfrozen thespian agent}: the same pre-trained thespian agent but with unfrozen weights and no attention mechanism.
\end{itemize}
Both agents are trained on a new ``Rogue'' reward, which rewards the agent for the union of thief-specific and adventurer-specific actions.

For the thespian attention agent, we measure the total ``rogue'' game score after each step. 
For the unfrozen agent, we measure the total ``rogue'' game score as well as just the thief score and just the adventurer score. 
Whereas the thespian attention agent is frozen and cannot lose its ability to emulate a thief or adventurer (character prompt and internal weights are unchanged), the unfrozen agent may lose its ability to emulate the thief and adventurer as it trains on the ``rogue'' reward. 

\subsection{Results}

Figure \ref{fig:Games} shows the total cumulative score for the thespian attention agent and unfrozen agent, averaged across five training runs each.
In all three maps, the thespian attention agent training a new ``rogue'' prompt outperforms the unfrozen agent training a new ``rogue'' prompt.
In the adventurer-first and alternating maps the thespian attention agent has converged by 1,500 steps.

The unfrozen agent training a new ``rogue'' prompt fails to converge within the allotted time. The unfrozen agent converges after 15,000 steps, which is 10x slower than the frozen thespian agent with attention mechanism, though it does match the performance eventually. However, we also see that the unfrozen agent quickly loses its ability to emulate the plain thief and plain adventurer. The unfrozen agent can be trained using a rotation of games for all three characters.
When this is done it takes in excess of 40,000 steps to before the it converges on a model that can play all three characters.

The reason the thespian attention agent does not do as well on the thief-first map as the others is because of bias introduced in the pre-training. Because the training regimen alternates characters, it trains on the ``thief'' character last. 
This makes the thespian agent slightly overfit to the thief character (relative to the adventurer).
While this might seem like it might give it an advantage on the thief-first map, in means that it takes longer to encounter non-thief ``rogue'' rewards; the encounter of early thief rewards reinforces this by placing more attention weight on thief action logits.

\section{Ablation Studies}

We investigate three alternative ways to incorporate attention into the thespian agent:

\begin{itemize}
\item Attention over a direct weighted average of character prompts. 
\item Attention over a weighted average of the soft character prompt plus state observation.
\item Attention over action probabilities instead of raw logits.
\end{itemize}
The first two, which focused on attention over the soft character prompts in various ways, 
resulted in agents that failed to learn a new character. 
The agent would choose actions that went with the most attended prompt and would never achieve blending.
This is because the attention layer would just act as a scalar on the inputs.

The third alternative would have used a softmax layer to convert action logits to a probability distribution before being fed into the attention mechanism.
In all cases, this variation was inferior to operating on raw logits.
The softmax conversion of raw logits to a probability distribution smooths the values, making it harder to discriminate between actions. 
Manipulating the logits allows for the biases of the individual character prompts to be more faithfully preserved. 

\section{Conclusions}

In this paper,  
we make the distinction between {\em character} agents and {\em actor} agents.
A character agent learns a model of a single character.
An actor, or thespian, agent learns a model of multiple characters and can take direction through a soft prompt about which character to emulate at any given time.
Our formulation of a thespian agent is further able to reason about which actions would be appropriate to each character.

The production of different action logit scores for different characters allows us to add an additional attention mechanism that learn new characters that remix previously known characters in a few-shot fashion. 
This is shown by training a new character that can take on the behavioral characteristics of previously known characters to respond to new circumstances in the environment. 

In the context of text role-playing games, a grand challenge for AI\cite{ccb2022dungeons}, 
this work presents a step toward open-ended agents with disentanglable behavior policies.

\clearpage
\bibliography{aaai23}

\clearpage
\appendix
\begin{figure*}
    \centering
\includegraphics[width=\linewidth]{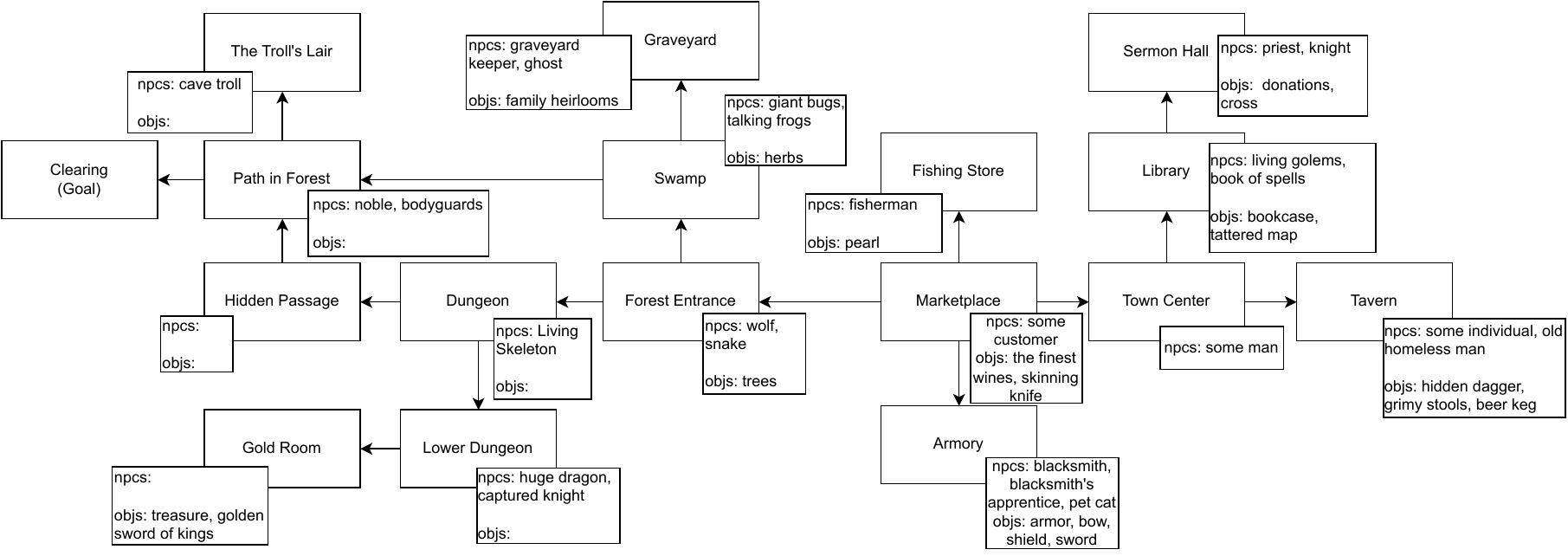}
    \caption{LIGHT Map}
    \label{fig:full-map}
\end{figure*}
\section{Appendix}
\subsection{LIGHT Map}
Figure \ref{fig:full-map} shows the entire LIGHT map layout used for experimentation.

\subsection{Training Details}
While most other hyperparameters are kept the same, we increase the learning rate while decreasing the value loss for the thespian attention. Despite the new prompt and the Attention Module having comparably a smaller number of trainable parameters, we also train over a much smaller number of steps to emulate Few-Shot training. Where thespian agent allowed to train to completion over 10,000 games, we constrain the thespian attention to only 3000 steps, which for a well performing agent could be potentially 150 games but could also potentially only be 40 games for a nonperforming agent, depending on the number of steps the agent takes within a game. While we found a higher learning rate hinders the thespian agent, for the thespian attention the higher learning rate benefited the agent due to the agent having already learned and being constrained to a smaller, more optimal set of actions. 

We also lower the coefficient of the value loss as well as changing how the value is calculated. As the Critic is frozen, we know it will always output the wrong reward value for any ``Adventurer" or ``Thief" action that isn't included in the new character. This results in large amounts of unnecessary loss that throws off the fusion agent during training. However, the value loss cannot be removed completely as it comprises the vast majority of the loss due to the pre-training of the thespian agent prior to the thespian attention.

\end{document}